\pdfoutput=1
\documentclass[conference]{IEEEtran}
\IEEEoverridecommandlockouts

\usepackage{cite}
\usepackage{amsmath,amssymb,amsfonts}
\usepackage{algorithmic}
\usepackage{graphicx}
\usepackage{textcomp}
\usepackage[most]{tcolorbox}
\usepackage{booktabs}
\usepackage{multirow}
\usepackage{siunitx}
\usepackage{caption}
\usepackage{subcaption}
\usepackage{tikz}
\usepackage{soul}
\usepackage{makecell}
\usepackage[hidelinks]{hyperref}

\usetikzlibrary{arrows.meta, positioning, shapes.geometric}

\usepackage{agentprompt}
\usepackage{iccad_userinput}
\usepackage{llmexplain}

\usepackage{fancyhdr}

\setulcolor{Red}

\newtcolorbox{llmoutput}{
    colback=black!5!white, 
    colframe=black!75!black, 
    fonttitle=\bfseries,
    boxsep=1mm,
    arc=1mm, 
    boxrule=0.5pt,
    fontupper=\itshape, 
    left=2mm,
    right=2mm,
    top=1mm,
    bottom=1mm
}

\def\BibTeX{{\rm B\kern-.05em{\sc i\kern-.025em b}\kern-.08em
    T\kern-.1667em\lower.7ex\hbox{E}\kern-.125emX}}
\begin{document}

\title{AnaFlow: Agentic LLM-based Workflow for Reasoning-Driven Explainable and Sample-Efficient Analog Circuit Sizing\\
}

\author{\IEEEauthorblockN{Mohsen Ahmadzadeh, Kaichang Chen, Georges Gielen}
\IEEEauthorblockA{\textit{ESAT-MICAS, KU Leuven, 3001 Leuven, Belgium} \\
\{mohsen.ahmadzadeh, kaichang.chen, georges.gielen\}@kuleuven.be}
}

\IEEEpubid{\begin{minipage}{\textwidth}\ \\\\\\\\[12pt] \footnotesize \textcopyright~2025 IEEE. This article has been accepted for publication by the 2025 International Conference on Computer-Aided Design (ICCAD 2025) and was presented in Munich, October 2025. All copyrights are reserved for IEEE. 
\end{minipage}}

\maketitle

\pagestyle{fancy}
\fancyhead{} 
\fancyfoot{} 
\renewcommand{\headrulewidth}{0pt}
\fancyhead[C]{\footnotesize Accepted for 2025 IEEE/ACM International Conference On Computer-Aided Design (ICCAD)}
\fancyfoot[C]{\thepage}

\begin{abstract}
Analog/mixed-signal circuits are key for interfacing electronics with the physical world. Their design, however, remains a largely handcrafted process, resulting in long and error-prone design cycles. While the recent rise of AI-based reinforcement learning and generative AI has created new techniques to automate this task, the need for many time-consuming simulations is a critical bottleneck hindering the overall efficiency. Furthermore, the lack of explainability of the resulting design solutions hampers widespread adoption of the tools. To address these issues, a novel agentic AI framework for sample-efficient and explainable analog circuit sizing is presented. It employs a multi-agent workflow where specialized Large Language Model (LLM)-based agents collaborate to interpret the circuit topology, to understand the design goals, and to iteratively refine the circuit’s design parameters towards the target goals with human-interpretable reasoning. The adaptive simulation strategy creates an intelligent control that yields a high sample efficiency. The AnaFlow framework is demonstrated for two circuits of varying complexity and is able to complete the sizing task fully automatically, differently from pure Bayesian optimization and reinforcement learning approaches. The system learns from its optimization history to avoid past mistakes and to accelerate convergence. The inherent explainability makes this a powerful tool for analog design space exploration and a new paradigm in analog EDA, where AI agents serve as transparent design assistants.
\end{abstract}


\section{Introduction}
Analog and mixed-signal circuits form the essential interface between the digital world of computation and the continuous signals of the physical world. Their role is critical in a vast array of applications. Despite their ubiquity, the design of these circuits, however, remains a formidable challenge. Unlike digital design, which benefits from extensive automation, analog circuit sizing is still largely a handcrafted process reliant on the intuition and experience of human designers. This manual approach results in long, costly, and error-prone design cycles, creating a significant bottleneck in the development of modern electronic systems \cite{Barros, MBTD3, AnaCraft}.

\begin{figure}
    \centering
    \includegraphics[width=\linewidth]{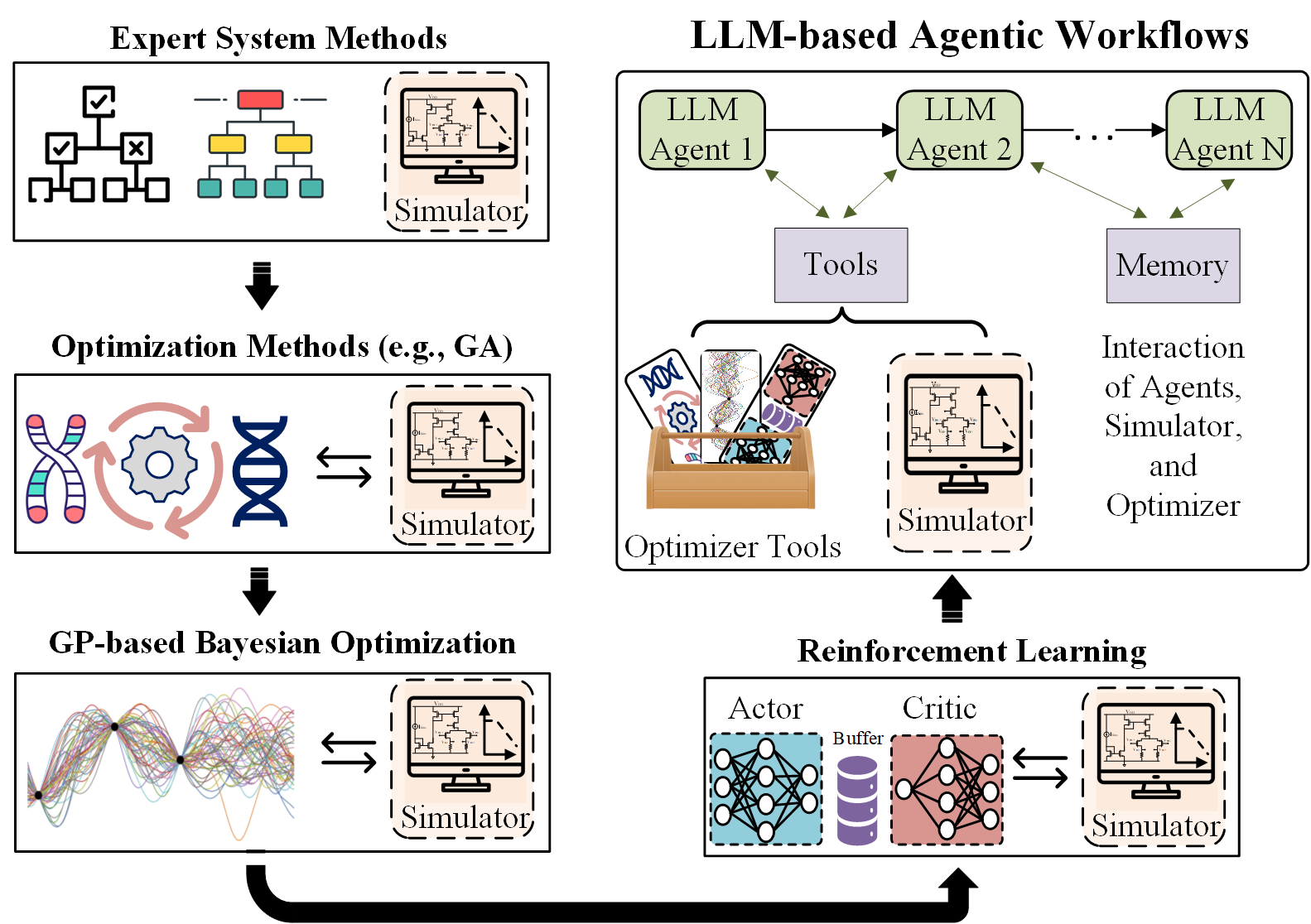}
    \caption{Historic trend of different approaches used for the automation of analog circuit sizing.}
    \label{fig:trend}
\end{figure}

To boost the automation process, the field of Electronic Design Automation (EDA) is increasingly turning to Artificial Intelligence (AI). Figure \ref{fig:trend} shows the history of approaches explored in research. It will be described in more detail in Section II. Early efforts employing expert systems, optimization methods such as Genetic Algorithms (GA) \cite{Wolfe, Pessoa} Bayesian optimization (BO) \cite{Batch, Gaspad}, and more recently Reinforcement Learning (RL) \cite{MBTD3, AnaCraft, L2DC, GCN-RL, AutoCkt, Prioritized, RobustAnalog, gm_id, Trust-Region, MATD3} have demonstrated the potential to automate the search for optimal circuit sizes. However, these pioneering techniques suffer from two fundamental limitations that hinder their widespread adoption. 
\begin{itemize}
    \item First, they are notoriously sample-inefficient, often requiring many thousands of computationally expensive circuit simulations to converge to a solution \cite{MBTD3, AnaCraft, AutoCkt}. This CPU time explosion problem makes them impractical for complex designs.
    \item Secondly, they function as "black boxes," providing solutions without offering any insight into them. This lack of explainability prevents designers from trusting the results, nor understanding the trade-offs used, leading to a critical lack of confidence in the automated solutions and the analog design automation tools that generate them.
\end{itemize}
To address the above challenges, we resort to the emerging Large Language Model (LLM)-based agentic AI paradigm \cite{LLM_Agents_1, LLM_Agents_2}. This paper introduces a novel framework, called AnaFlow, for analog circuit sizing that is both reasoning-driven, sample-efficient, and inherently explainable. Our AnaFlow approach moves beyond pure numerical optimization and mimics the cognitive workflow of an expert analog designer. We employ an agentic system where multiple specialized LLM agents collaborate to interpret connections, find matching and other constraints, formulate design steps based on their acquired analog circuit theory knowledge, execute targeted simulations, and iteratively refine the design towards the target design goals through reasoning feedback and through the calling of optimizers (RL, BO, etc.). By grounding the optimization process in an explicit, human-interpretable reasoning framework, AnaFlow has the potential to reduce the number of required simulations, to fully automate the sizing task, and to produce a fully justified design rationale besides finding optimized design parameters.

This paper makes the following key contributions:
\begin{itemize}
\item We present a novel, multi-agent LLM workflow that mimics the workflow of expert designers and brings structured reasoning and explainability to the analog circuit sizing task.
\item We demonstrate that the reasoning-driven approach achieves better sample efficiency, can call optimizers and can achieve target specifications using a fully automated process that requires minimal input from the users, as opposed to other established black-box optimization approaches.
\item We demonstrate our framework’s explainability by showing its capability in producing human-interpretable reasoning for the steps taken during the circuit sizing.
\item We show our framework's capability by sizing two operation amplifiers of different complexities; all through reasoning, planning, tool use and decision feedback within AnaFlow.
\end{itemize}

This paper is organized as follows. Section II will briefly review the evolution of automated analog circuit sizing approaches, from traditional optimization algorithms to recent LLM-based approaches, highlighting the remaining challenges. Section III will detail the architecture of our proposed AnaFlow agentic LLM workflow, describing the roles and interactions of the specialized agents incorporated. Section IV will present our experimental results, providing a quantitative analysis of our method's sample efficiency and a qualitative demonstration of its explainability. Finally, Section V will conclude the paper with a summary of our contributions.

\section{Background and Related Work}

Automating analog circuit sizing has been a long-standing pursuit in the Electronic Design Automation (EDA) community. The approaches have evolved significantly over the years, moving from expert systems over optimization methods to more machine learning-based solutions. These were depicted in Figure \ref{fig:trend}.

\subsection{Expert Systems and Optimization Methods}

Early attempts to automate analog sizing were based on expert system techniques \cite{Gielen_Rutenbar, BLADES}. They failed as it was difficult to make all designer expertise explicit.

Then came optimization methods that explore the vast design space by successively evolving candidate solutions. Example algorithms are simulated annealing and population-based genetic algorithms (GAs) \cite{Wolfe, Pessoa}. While capable of finding pretty optimal solutions, these methods are notoriously inefficient, often requiring many tens of thousands of circuit simulations to converge. Furthermore, their inherent randomness and lack of learning mechanisms mean that knowledge from one optimization run cannot be transferred to another one \cite{AutoCkt}. They also provide no insight into the design solutions obtained and the design choices made. 

Because of the computational explosion, Bayesian Optimization (BO) was adopted by researchers for its greater sample efficiency \cite{Batch, Gaspad}. The BO methods rely on Gaussian Processes (GPs) to build surrogate models of the objective function. This makes them more suited for expensive black-box problems. BO's applicability, however, is limited by its computational complexity, which scales cubically with the number of observations \cite{Bobak}. This excessive runtime makes it impractical for problems with more than 20 variables \cite{Tutorial}, rendering it unsuitable for complex circuits of truly high-dimensional design space \cite{MBTD3, AnaCraft}.

\subsection{Reinforcement Learning Approaches}
In recent years, Reinforcement Learning (RL) has emerged as a powerful paradigm for analog circuit sizing. Actor-Critic RL agents learn a sizing policy (the Actor neural network) by interacting with a circuit simulator (the environment), receiving rewards based on the performance metrics and maximizing the sum of the expected rewards (the Critic neural network) \cite{L2DC, GCN-RL, AutoCkt, Prioritized, RobustAnalog, gm_id, Trust-Region}. Recent studies have shown that RL can outperform both GA and BO in terms of sample efficiency and scalability. The learning-based nature of RL allows for transfer learning \cite{GCN-RL}, where a policy trained on one circuit can be fine-tuned for another one. Also, multi-agent schemes \cite{MATD3, MBTD3, AnaCraft} have demonstrated the potential to scale to complex, industrial-scale designs.
Despite these advances, RL-based methods still face significant hurdles. They still require hundreds to thousands of simulations to train an effective policy. More critically, the learned policy is still a \textit{black box}, typically encoded in the weights of the trained Actor-Critic networks. This opacity makes their decision-making process inscrutable, offering no explanation to the designer for why certain design parameter values were chosen. This lack of explainability is a major barrier to the approach's adoption in the industry, as human designers are hesitant to trust solutions they cannot interpret nor understand.

\subsection{The Rise of LLM-based Agentic AI}

Since the advent of Large Language Models (LLMs), a new paradigm for design automation is emerging, centered on LLM-based agentic workflows. LLM agents can leverage their vast pre-trained knowledge to reason, plan, and use tools to accomplish complex tasks, such as customer responding, automated software development, complex web navigation, social behavior simulation \cite{LLM_Agents_1, LLM_Agents_2}, etc. The power of these workflows stems not only from simple \textit{prompt engineering}, but also from sophisticated \textit{context engineering} \cite{context}, where the agents are dynamically provided with the necessary information—such as documents, data, and access to external tools—to execute a task. This ability to dynamically provide the right information and tools, in the right format, and at the right time in the flow is the core of context engineering, enabling dynamic and flexible problem solving \cite{context}.

The agentic paradigm is now also finding ground in the complex domain of EDA, where researchers are applying LLMs across the entire chip design stack. Also in analog design, generative AI and agentic workflows are emerging to tackle the remaining challenges. AnalogGenie \cite{AnalogGenie} focuses on the creative task of topology generation by training a GPT neural network on connecting devices and sub-blocks to produce analog circuit structures. AnalogCoder generates the circuit implementation code in PySpice through a feedback-enhanced, training-free flow \cite{AnalogCoder}. Beyond this topological design, agents are also being integrated into the optimization loops: ADO-LLM \cite{ADO-LLM} combines an LLM with Bayesian Optimization for efficient analog circuit sizing, leveraging the LLM's knowledge to guide the search for optimal parameters. The LLM acts as a knowledge-infused proposer in parallel with a Bayesian optimizer. However, it decides its own proposals rather than directing the overall workflow or providing a retrospective analysis of the full optimization trajectory and trade-offs, which is the focus of our AnaFlow framework in this paper.

\begin{figure}
    \centering
    \includegraphics[width=\linewidth]{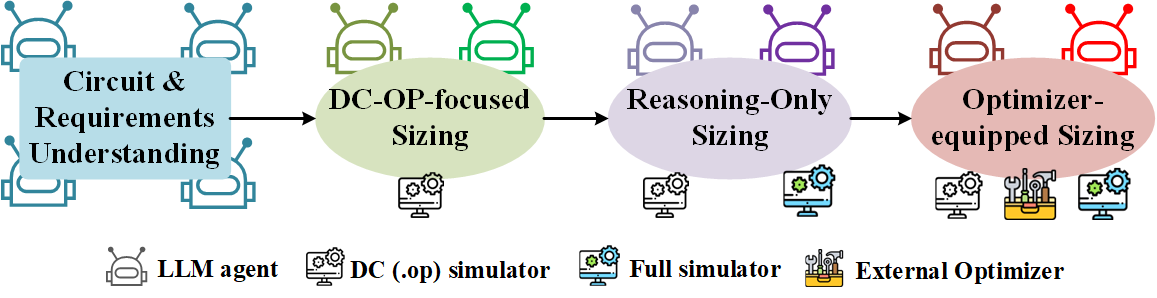}
    \caption{Overview of the four phases in AnaFlow.}
    \label{fig:FlowGraph}
\end{figure}

\begin{figure*}
    \centering
    \includegraphics[width=\linewidth]{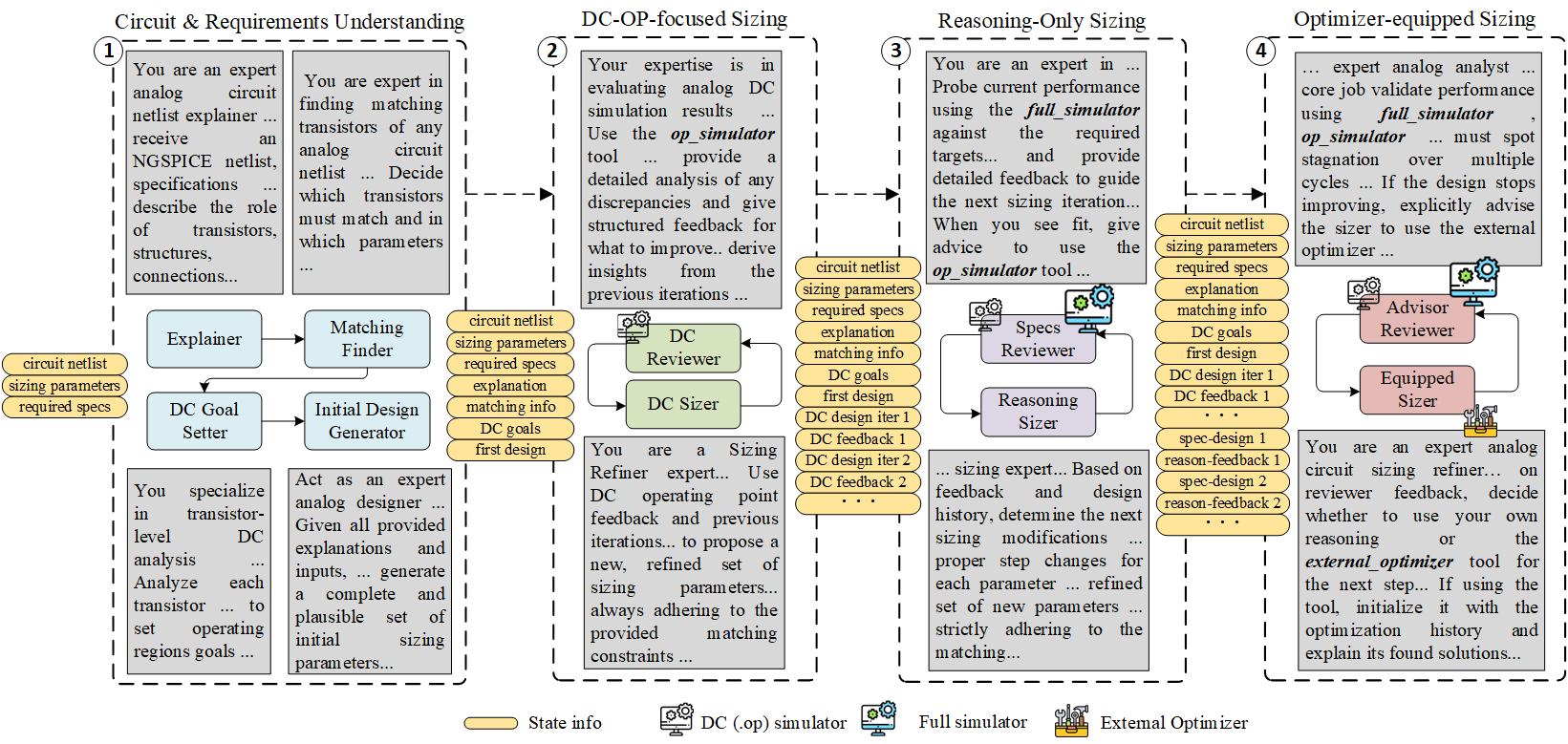}
    \caption{Detailed diagram of all agents in the AnaFlow workflow with the four phases of Figure \ref{fig:FlowGraph}.}
    \label{fig:workflow}
\end{figure*}

\section{Methodology}

\subsection{Foundations of the Agentic Design Workflow}
The efficacy of our agentic system is rooted in core LLM practices and capabilities: context engineering, \textit{tool integration}, and \textit{structured output} generation. Modern LLM applications depend critically on context engineering, where the input prompt is carefully crafted with instruction prompts and relevant data to guide the model toward a desired outcome. This goes beyond simple queries and represents the primary mechanism for specializing a general-purpose model towards a specific task. A pivotal breakthrough in this direction was to enable LLMs to use external \textit{tools}, as demonstrated by seminal works like Toolformer \cite{Toolformer}, LaMDA \cite{LaMDA} and ReAct \cite{ReAct}. This allows an LLM to overcome its inherent knowledge limitations by interacting with external software tools, such as invoking a simulator, retrieving data from a database, or performing a web search. By integrating an LLM with tools, it can ground its reasoning in real-world data and act upon its environment, forming the basis of an autonomous agent. Our work leverages these foundational principles to construct a specialized agentic workflow tailored specifically for the task of analog circuit sizing.

\subsection{The AnaFlow Agentic Workflow}
In the AnaFlow agentic framework, specialized LLM agents collaborate to mimic the structured reasoning process of an expert analog designer. The complex sizing task is broken up in a carefully orchestrated sequence of four phases (see Figure \ref{fig:FlowGraph}): an initial analysis phase to understand the circuit, followed by three distinct, iterative loops for DC-operating-point-(DC-OP)-focused sizing, reasoning-only sizing, and optimizer-equipped sizing. The latter can call external optimizers like BO or RL. This modular architecture ensures a reliable and explainable automated path to optimized solutions. The details of the overall workflow of Figure \ref{fig:FlowGraph} will be discussed below (see Figure \ref{fig:workflow}).

\subsection*{}
\textit{B.1. Understanding the Circuit and Requirements}

The workflow begins with a deep analysis of the sizing problem, handled by a sequence of specialist agents that deconstruct the task into manageable parts. Figure \ref{fig:workflow}.1 shows the context provided to each agent in this first phase:

\begin{itemize}
    \item \textbf{Circuit Explainer}: this agent analyzes the circuit topology given in the input SPICE netlist. It reasons about all the connections of devices in the netlist, tries to find different structures in the circuit and provide explanations about the role of each device or structure in the topology.
    \item \textbf{Matching Finder:} this agent identifies devices in the circuit that require precise matching or symmetry for optimal performance, such as differential pairs or current mirrors. It annotates these components, allowing the sizing agents to apply constraints that preserve these critical requirements during optimization.
    \item \textbf{DC Goal Setter}: this agent reads the information gathered from the previous agents, reasons about them, and finds the DC and operating-region requirements that each transistor of the circuit needs to work in.
    \item \textbf{Initial Designer}: this agent represents the initial sizing step. It synthesizes the information from the previous agents and, based on the inherent knowledge of the LLM about analog circuits, generates a plausible initial set of device sizes that it think leads to meeting the required specifications.
    
\end{itemize}

\subsection*{}
\noindent \textit{B.2. DC-OP-focused Sizing}

Before committing to expensive performance simulations, the workflow enters a preparatory second phase (see Figure \ref{fig:FlowGraph}) to establish a reasonable DC biased solution. This is handled by a specialized iterative loop, intentionally constrained to a few cycles ($\sim$5-10) to limit LLM calls and runtime. The objective here is not to reach perfect DC convergence, but rather to rapidly gather insights into the circuit's behavior in the target technology; much like a human expert's initial steps. Figure \ref{fig:workflow}.2 shows the context of the agents in this phase:
\begin{itemize}
    \item \textbf{DC Reviewer}: This agent takes care of the DC validation process. It takes the current set of design parameters, performs a fast DC operating point (\verb|.op|) SPICE simulation, and parses the simulation output to understand the operating region of each transistor. The agent then compares these simulated results against the target DC goals set before, identifying discrepancies and suggesting reasoning-based overall modifications for resolving them.
    \item \textbf{DC Sizer}: this agent receives the discrepancy report and the modification feedback from the \textbf{DC Reviewer} and, based on the inherent knowledge of the LLM about analog circuit behavior, determines the specific parameter modifications to resolve the identified issues. Its output is a \textit{structured output} of parameter names and their new refined values, which are fed back to the \textbf{DC Reviewer} for the next validation cycle.
\end{itemize}

\subsection*{}
\noindent \textit{B.3. Reasoning-Only Sizing}

After a few iterations of DC refinement, the workflow transitions to its first full performance optimization loop: checking all required specifications through full simulation. This third phase relies purely on the LLM's inherent, learned knowledge of analog circuit theory combined with the context gathered by the previous agents to refine the design—no external numerical optimizers are used yet in this phase. This stage emulates an experienced designer applying his/her intuition and making heuristic-driven adjustments. It is a sample-efficient loop, designed to quickly find the best possible design that can be found using solely the agent's ability to reason about complex circuit trade-offs before resorting to simulation-expensive optimizers (BO, RL). We limit this loop to a maximum of 20 full simulations. Figure \ref{fig:workflow}.3 shows the context provided to the agents in this phase:
\begin{itemize}
\item \textbf{Specs Reviewer}: This agent's job is to carry out a full performance validation and provide reasoning-based feedback for modifications to improve the performance metrics. It takes the latest sizing parameters and employs a comprehensive simulation \textit{tool} that executes more complex analyses (e.g., \verb|.ac|, \verb|.tran|) to measure key performance metrics like voltage gain, phase margin, unity-gain bandwidth, and power consumption. The agent's output is a combination of critique and feedback, identifying the most significant performance deviations to be addressed, and providing insights and reasoning on which changes to design parameters need to be made for improving those performance metrics. This agent also has access to DC-only simulation tool; it can call this when it feels that the sizing is in a very bad condition or when it is confused and needs cheap DC simulation insights. The choice of which tool to use when is based on the agent's own judgment and reasoning.
\item \textbf{Reasoning Sizer}: Acting on the feedback from the \textbf{Specs Reviewer} and based on the inherent analog circuit knowledge of the LLM as well as the optimization history, this agent provides specific parameter modifications, delivered as a \textit{structured output} of new design parameter values for the next simulation cycle.
\end{itemize}

\subsection*{}
\vspace{-5pt}
\noindent \textit{B.4. Optimizer-equipped Sizing}

When the pure reasoning-based loop in the third phase reaches its iteration limit, the workflow escalates to the fourth and final optimization stage. Here, the agents can call external numerical optimizers (e.g., BO or RL) to handle particularly challenging trade-offs or to overcome local minima that defy heuristic adjustments. Figure \ref{fig:workflow}.4 shows the context of the two agents used here:

\begin{itemize}
\item \textbf{Advisor Reviewer}: This agent performs the same core functions as in the reasoning-only phase, including full performance validation and providing detailed critiques of metric deviations and feedback for sizing adjustment. It additionally monitors the optimization progress for signs of stagnation, such as repeated patterns of diminishing returns or failure to improve key metrics over several cycles. When such conditions are detected, the agent explicitly advises to invoke an external optimizer, reasoning about why it is needed. It retains access to the DC-only simulation \textit{tool} for quick checks.
\item \textbf{Equipped Sizer}: Building directly on the feedback and any optimizer advice from the \textbf{Advisor Reviewer}, this agent evaluates whether to rely solely on its own LLM-based reasoning for parameter refinements—much like the Reasoning-Only Sizer—or to activate an external optimizer \textit{tool} for a more systematic search. When opting for the optimizer, it configures the tool with initial points (from the previous simulations and the optimization history) and decides the simulation budget for that run, and finally integrates the optimizer's suggestion into a \textit{structured output} of refined parameters. 

\end{itemize}

\begin{figure}
    \centering
    \includegraphics[width=1\linewidth]{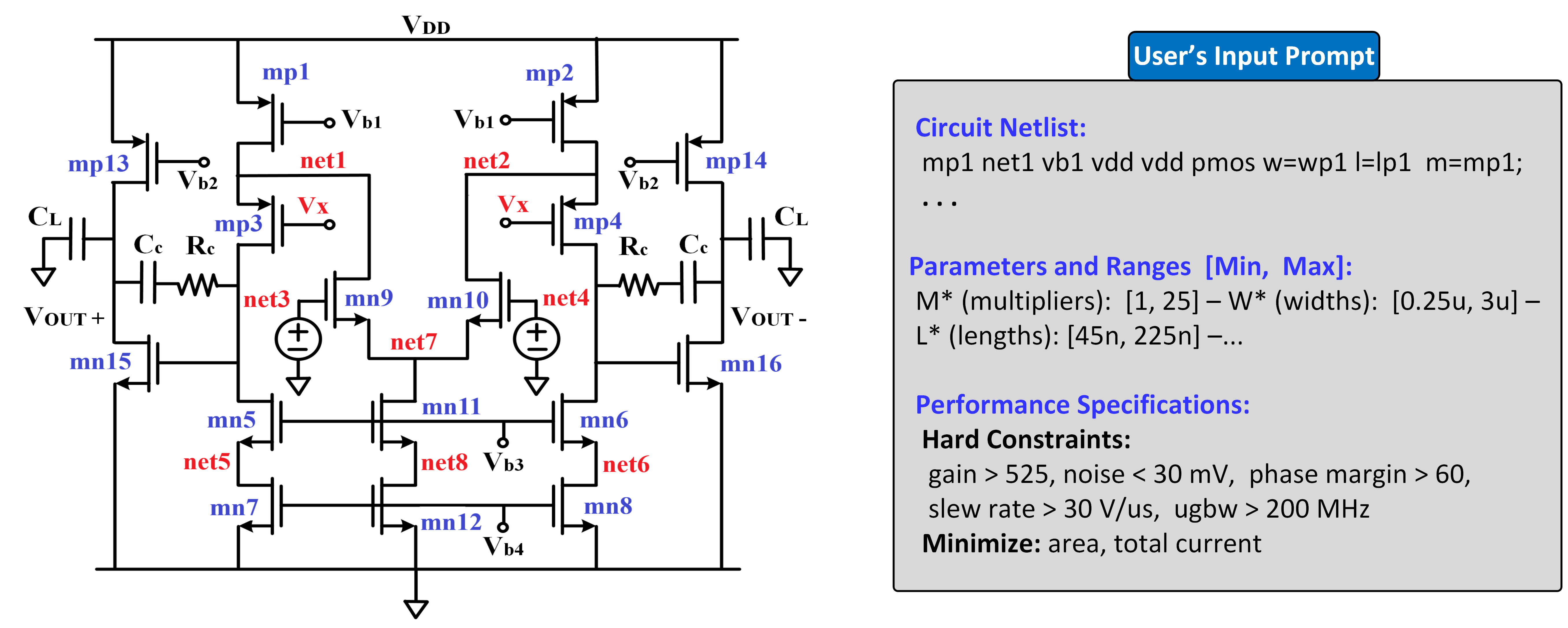}
    \caption{Schematic of the fully differential folded-cascode opamp (its CMFB is not shown), and user input for sizing the circuit fully automatically in AnaFlow.}
    \label{fig:user_input}
\end{figure}

\section{Experimental Results}

The AnaFlow agentic workflow is now demonstrated to size two circuits with increasing levels of complexity: a two-stage Miller opamp and a differential folded-cascode opamp with common-mode feedback (CMFB).

Since we are using LLMs that run their inference in the cloud, we use the open-source predictive BSIM 45nm PDK \cite{45nm} with NGSPICE \cite{Ngspice} as simulator. Figure \ref{fig:user_input} shows the schematic of the differential folded-cascode opamp and all the entire information about this circuit given as input to the framework, namely: the netlist, the sizing parameters and their ranges, the required performance metrics, and the optimization targets. As can be seen, minimal input needs to be given to the workflow.

Our AnaFlow agentic framework is implemented in Google's ADK \cite{ADK} which provides sequential agents, loop agents, the use of tools, and state (memory) management capabilities. It also provides the ability to use LLMs other than Gemini, such as GPT, Claude, Grok, etc. 

Below, we first provide the sample-efficiency results by comparing our method to the previously published RL-based analog sizing optimizer AnaCraft \cite{AnaCraft, MBTD3}. Then, we show examples of explanations and decisions made by the agents throughout the steps of the sizing workflow, to demonstrate the explainability capabilities of our framework.

\subsection{Sample-Efficiency and Runtime Comparison}

As the LLM agents in our framework guide the sizing process through reasoning, it considerably requires less simulations; simulations are either called by the LLM or through the use of the external optimizer. The better the LLM agent (the larger and smarter the model), the better it is in reasoning and the more circuit knowledge it inherently has, leading to even fewer external optimizer calls and less full simulations in total.

On the other hand, smarter and larger LLM models take more time to respond to each call. Therefore, although considerable sample efficiency is gained by larger LLMs, they reversely may lead to longer runtimes. This drawback is less prominent for circuits with much longer full simulation runtimes.

The results are shown in Table \ref{tab:llm_performance_comparison}. As can be seen, for the two-stage opamp, Gemini 2.5 Pro can find a sizing solution that satisfies all hard performance constraints in the specifications within 10 iterations of full simulations and reasoning, without having to call the external optimizer. For the more complex differential folded-cascode opamp, it can find a solution after only 1 call to the external optimizer —the single-agent version of the RL-based AnaCraft algorithm \cite{AnaCraft, MBTD3}— and still requires to a total of less than only 100 simulations. However, these results do come at the price of a longer runtime and a higher cost per token. Faster LLM models like Gemini 2.5 Flash, Claude Sonnet 4, and Grok 3 require more reasoning steps, more calls to the external optimizer, and therefore, a higher number of simulations, but their overall runtime is better, as evidenced in Table \ref{tab:llm_performance_comparison}. 

We experienced two main problems with some of the LLM models used in our experiments; 1) complying to a long structured output for a simulation call, and 2) very long runtimes. For example, the first problem caused some models to fail for the more complex folded-cascode opamp as the circuit has many more parameters and therefore a longer and more sophisticated structured output which those models failed to provide at each call to the simulator. On the other hand, some larger models needed significantly longer runtimes while not bringing any sample-efficiency benefits, whlie others had such a long runtime that they could not reach any significant step within 2 hours while having a very high cost per token.

\begin{figure}[t!]
    \centering
    \includegraphics[width=0.83\linewidth]{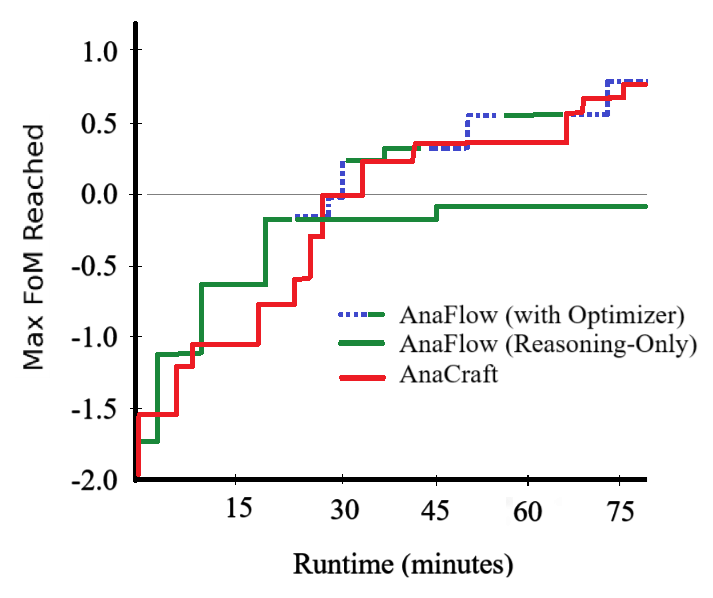}
    \caption{Runtime comparison of AnaFlow and AnaCraft \cite{AnaCraft, MBTD3} in reaching optimized design solutions.}
    \label{fig:fom}
\end{figure}

Figure \ref{fig:fom} shows a runtime comparison of the LLM-based AnaFlow framework (with Gemini 2.5 Pro as the LLM) and the single-agent AnaCraft RL algorithm of \cite{AnaCraft, MBTD3} in reaching an optimized solution with a Figure of Merit (FoM) of above 0.5 for the differential folded-cascode opamp with CMFB (the FoM is calculated according to \cite{MBTD3, AnaCraft}). It can be seen that LLM-based workflows like AnaFlow today still need similar runtimes as the state-of-the-art RL-based algorithms.

\begin{table}[t]
\caption{Comparing the agentic workflow (using different LLMs) to RL based on finding the first sizing solution.}
\label{tab:llm_performance_comparison}
\begin{center}
\begingroup
\setlength{\tabcolsep}{3pt} 
\footnotesize 
\begin{tabular}{|c|c|c|c|c|c|c|c|c|}
\hline
\multirow{2}{*}{\textbf{Circuit}} & \multirow{2}{*}{\textbf{Approach}} & \multirow{2}{*}{\textbf{\shortstack{LLM \\ Calls}}} & \multirow{2}{*}{\textbf{\shortstack{Opt. \\ Calls}}} & \multirow{2}{*}{\textbf{\shortstack{DC\\ Sims.}}} & \multicolumn{3}{c|}{\textbf{Full Simulations}} & \multirow{2}{*}{\textbf{\shortstack{Time \\ (min)}}} \\
\cline{6-8}
& & & & & \textbf{LLM} & \textbf{Opt.} & \textbf{Total} & \\ 
\hline
\multirow{7}{*}{\makecell{Two-\\Stage\\Amp.}} 
 & Gemini 2.5 Flash & 81 & 5 & 5 & 32 & 250 & 282 & 16  \\
 & Gemini 2.5 Pro & 34 & 0 & 6 & 9 & 0 & 9 & 24 \\
 & GPT o3 & 68 & 4 & 7 & 25 & 180 & 205 & 46 \\
 & Claude Sonnet 4 & 52 & 6 & 24 & 19 & 300 & 319 & 17 \\
 & Claude Opus 4 & 36 & 1 & 16 & 11 & 50 & 61 & 27  \\
 & Grok 3 & 70 & 7 & 8 & 28 & 270 & 298 & 18  \\
 \cline{2-9}
 & AnaCraft \cite{AnaCraft, MBTD3} & N/A & 1 & N/A & N/A & 1035 & 1035 & 14 \\
\hline
\multirow{5}{*}{\makecell{Diff.\\Folded\\Cascode\\Amp.}} 
 & Gemini 2.5 Flash & 50 & 8 & 9 & 18 & 300 & 318 & 17  \\
 & Gemini 2.5 Pro & 42 & 1 & 6 & 14 & 50 & 64 & 29  \\
 & GPT o3 & 110 & 8 & 23 & 48 & 312 & 360 & 82 \\
 & Grok 3 & 116 & 7 & 15 & 46 & 280 & 316 & 22  \\
 \cline{2-9}
 & AnaCraft \cite{AnaCraft, MBTD3} & N/A & 1 & N/A & N/A & 1808 & 1808 & 31  \\
\hline
\end{tabular}
\endgroup
\end{center}
\end{table}

\subsection{Explainability of the Workflow}

\begin{figure}[b!]
\centering
    \includegraphics[width=\linewidth]{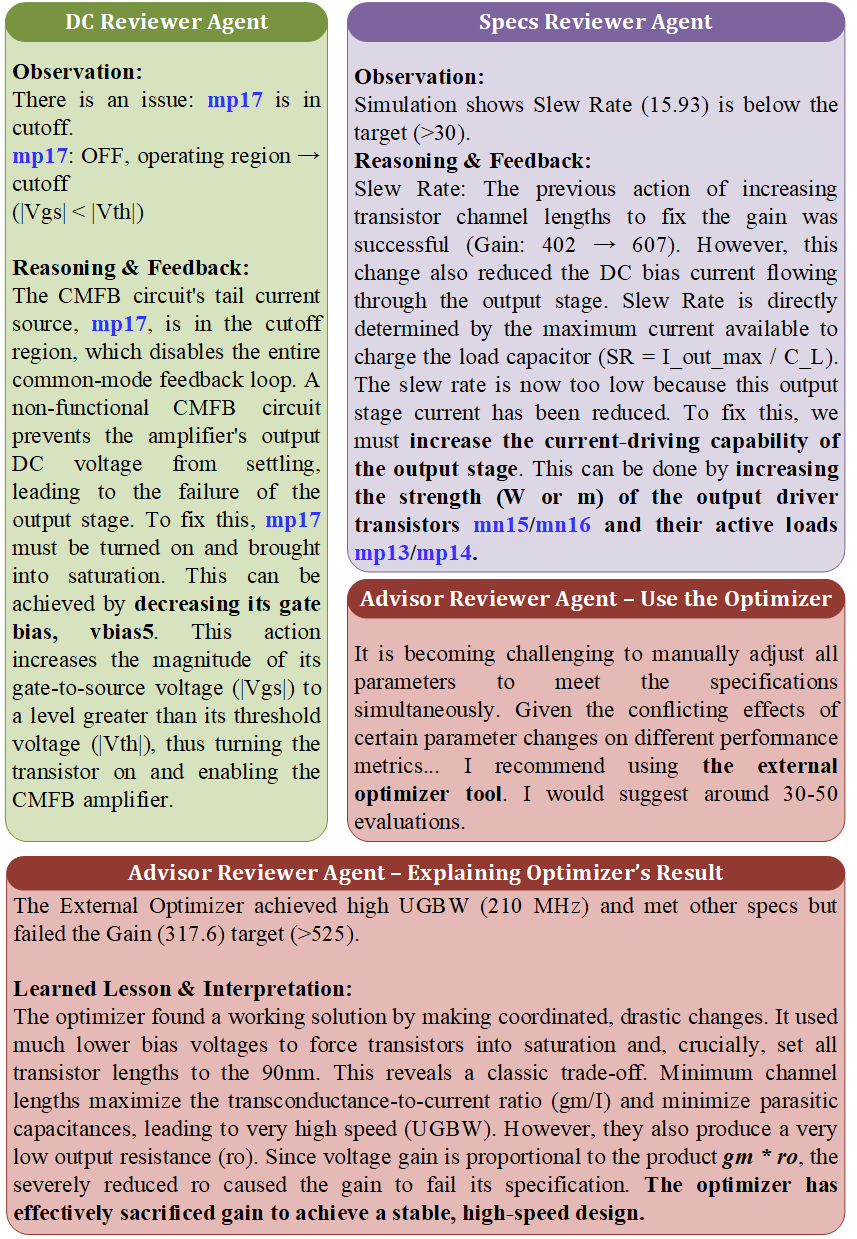}
    \caption{Illustrative examples of the LLM agents' explanations in direct reasoning and problem diagnosis.}
    \label{fig:llm_explain_direct}
\end{figure}

In order to demonstrate the explainability of the decisions made by our agentic workflow, we show some examples of reasoning and explanation provided by the LLM agents in Figure \ref{fig:llm_explain_direct}. As shown, the LLM (Gemini 2.5 Pro in these figures) can explain for example: 
what causes a DC failure and which design parameter to change for fixing it;
how do changes to channel lengths cause changes in the Gain and Slew Rate, and which sizing parameters need to change to achieve the target goals for those performance metrics. It can also decide that its own reasoning (which it calls manual adjustment) is stuck and that it advises to run the external optimizer. It can even learn lessons from the output of the optimizer and try to make sense of it after observing its actions and the corresponding outcomes (through reasoning based on the LLM's inherent knowledge of analog circuits). The workflow's answers provide insightful trade-off analysis and can explain the balance struck between conflicting performance metrics like gain, stability, and speed. 

This capability is unprecedented and could not be seen in any of the analog sizing optimization tools published before. Hence, using the AnaFlow framework, designers can better understand the sizing choices and get more insight from the automated sizing process. Designers have access to traceable design decisions, can learn from them, or criticize them. 


\section{Conclusions}
This paper has introduced the AnaFlow agentic LLM workflow for reasoning-driven, sample-efficient, and explainable analog circuit sizing. The framework employs a collaborative multi-agent LLM-based workflow that mimics the cognitive process of expert human designers. 
The structured approach, which decomposes the sizing task into analysis, DC refinement, and heuristic and optimization-based refinement steps, brings human-interpretable reasoning into the design process. The experimental results have shown a large reduction in the number of simulations required compared to straightforward optimization.
The framework's key contribution is its inherent explainability, providing transparent justifications for the design choices and fostering trust in the generated solutions. This work represents a significant step towards a new paradigm in analog EDA, where AI agents act not as opaque optimizers, but as transparent design assistants that accelerate the design cycle while augmenting the designer's own understanding.

\section*{Acknowledgment}
This work has been funded by the ERC Advanced Grant AnalogCreate (n° 101019982).


\end{document}